\documentclass[10pt,twocolumn,letterpaper]{article}

\usepackage{cvpr}              % To produce the CAMERA-READY version
\usepackage[dvipsnames]{xcolor}

\definecolor{cvprblue}{rgb}{0.21,0.49,0.74}
\usepackage[pagebackref,breaklinks,colorlinks,citecolor=cvprblue]{hyperref}

\usepackage[accsupp]{axessibility}
\usepackage{algorithm}
\usepackage{algorithmic} % 为了加算法
\usepackage{multirow}

\def\paperID{2967} % *** Enter the Paper ID here
\def\confName{CVPR}
\def\confYear{2024}

\title{StreamingFlow: Streaming Occupancy Forecasting with Asynchronous Multi-modal Data Streams via Neural Ordinary Differential Equation}

\author{Yining Shi$^{1,3}$\thanks{Work done during internship at DiDi Chuxing. \newline \ \ \ \ \ \indent \ $\dagger$  Corresponding author: Kun Jiang, Diange Yang.} \ \  Kun Jiang$^{1\dagger}$ \  Ke Wang$^{2}$ \ Jiusi Li$^{1}$ \ Yunlong Wang$^{1}$ \ Mengmeng Yang$^{1}$ \ Diange Yang$^{1\dagger}$\\
$^{1}$~School of Vehicle and Mobility, Tsinghua University \ \ \  $^{2}$~KargoBot, Inc $^{3}$~DiDi Chuxing
}

\begin{document}
\maketitle

\begin{abstract}
Predicting the future occupancy states of the surrounding environment is a vital task for autonomous driving. However, current best-performing single-modality methods or multi-modality fusion perception methods are only able to predict uniform snapshots of future occupancy states and require strictly synchronized sensory data for sensor fusion. We propose a novel framework, StreamingFlow, to lift these strong limitations. StreamingFlow is a novel BEV occupancy predictor that ingests asynchronous multi-sensor data streams for fusion and performs streaming forecasting of the future occupancy map at any future timestamps. By integrating neural ordinary differential equations (N-ODE) into recurrent neural networks, StreamingFlow learns derivatives of BEV features over temporal horizons, updates the implicit sensor's BEV features as part of the fusion process, and propagates BEV states to the desired future time point. It shows good zero-shot generalization ability of prediction, reflected in the interpolation of the observed prediction time horizon and the reasonable inference of the unseen farther future period. Extensive experiments on two large-scale datasets,nuScenes~\cite{nuScenes} and Lyft L5~\cite{Lyft},  demonstrate that StreamingFlow significantly outperforms previous vision-based, LiDAR-based methods, and shows superior performance compared to state-of-the-art fusion-based methods. 
%Occupancy represents the most basic and essential attribute of a driving scene. With the development of mapping raw data to bird's eye view (BEV), semantic occupancy forecasting is starting to become an important supplement for the environmental perception and prediction of autonomous driving. However, existing methods conduct occupancy forecasting in a way that only predicts uniform snapshots of future occupancy states, as they are trained to do, given a uniform data stream or synchronized multi-modal data stream as input. Since strict synchronization and discrete uniform prediction may not be representative and useful for real-world vehicles. We'd like to ease these two assumptions. 

\end{abstract}    
\section{Introduction}\label{sec:intro}
Occupancy grid is gaining more traction in the self-driving community, due to its versatilities in downstream tasks, e.g., irregularly shaped object representation, robot navigation, etc.
With the help of modern deep learning technologies, occupancy grid maps (OGMs) have evolved significantly. For example, not just LiDAR point clouds, camera images can now be used to construct high-quality occupancy grid maps~\cite{LSS}. 
In addition to its fine-grained geometric information, occupancy grids can also represent rich semantic and motion cues~\cite{FIERY,ST-P3,StretchBEV,UniAD}. Such cues grant occupancy networks the ability to predict the occupancy state of the environment at desired future timestamps. While trajectory forecasting only contains object-level information, the occupancy grid conveys a much more comprehensive understanding of the surrounding environment, which is the key to achieving better intelligence and driving safety.

\begin{figure}[!t]
	\centering
\includegraphics[width=0.45\textwidth]{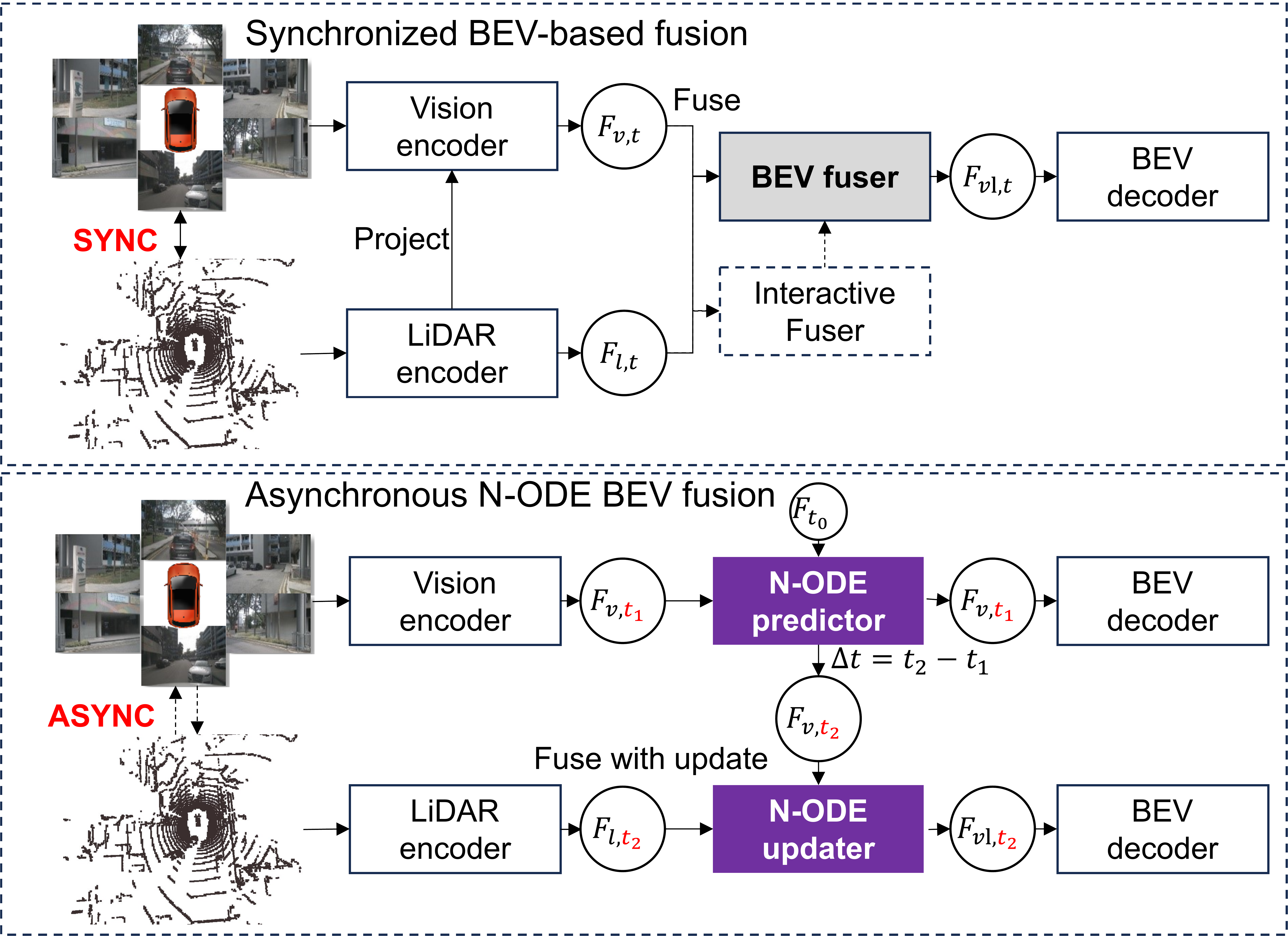}
	\caption{Comparison between conventional synchronized BEV fusion (top) and our asynchronous BEV fusion (bottom). We formulate fusion with an update-predict-update approach.}
	\label{comparison_fusion}
 \vspace{-1em}
\end{figure}

Despite the great potential of occupancy-based prediction, we notice that existing methods can only predict at fixed frequencies~\cite{FIERY,ST-P3,StretchBEV,UniAD, FusionAD}. This comes from the fact that sensors sample the environment at predetermined frequencies, e.g., motors in mechanical LiDARs spin at 10Hz, and many cameras captures images at 30FPS. Thus, previous algorithms predict at the same intervals as they are trained with ground-truth labels of future intervals.
If future occupancy can be predicted at any given timestamps in a continuous manner, self-driving algorithms can have shorter latency and more rapid responses.
Streaming forecasting can also relax the sensor synchronization constraints, as the continuous forecasting model can update the prediction as new sensory data come in.
Compared with discrete snapshot-based prediction, streaming motion forecasting~\cite{streamforecast} naturally has safety advantages.

We noticed several challenges in fulfilling such a streaming forecasting paradigm. Firstly, sensory data and ground truth annotations are collected at a fixed frequency. 
Streaming forecasting aims to achieve temporally dense and continuous prediction, with data only sparsely and uniformly sampled in the temporal domain.
Secondly, mainstream network architectures are not designed for such streaming requirements. Forecasting at any given timestamp requires delicate modeling of the temporal dynamics, but how to embed the temporal dynamics into the widely adopted BEV feature representation remains under-explored.

To this end, we propose to use the neural ordinary differential equation (N-ODE) framework for such temporal dynamics modeling. N-ODE is originally proposed for learning continuous implicit processes in dynamic systems. % It is widely adopted for time series analysis. Most relevantly, N-ODE is used for modeling GPS trajectory prediction\cite{node_predict}.
%\textcolor{red}{Here we need to describe some properties of N-ODE, and what people use it to do before we use it for streaming occupancy.}
% a novel StreamingFlow method for streaming occupancy forecasting. Our method leverages a neural ordinary differential equation (N-ODE) framework to propagate BEV features temporally.
In our application, by representing the continuous motion trends of the driving scenarios into a time series of implicit spatial BEV features, the neural ODE framework not only allows streaming occupancy predictions on any temporal horizon with one fixed model, but also relaxes the sensor synchronization constraints.

With such flexibility, we propose a novel  method, StreamingFlow, that performs asynchronous multi-modal sensor fusion to achieve streaming occupancy forecasting. Toward the goal of streaming modeling of temporal BEV features, StreamingFlow integrates neural ordinary differential equations (N-ODE) onto wrapped-based recurrent neural networks. It learns derivatives of BEV features over temporal horizons, updates the implicit sensor's BEV feature as part of the fusion process, and propagates BEV states to the desired future time point. A graphical illustration of fusion strategies between conventional BEV-based fusion and ours is illustrated in Fig.~\ref{comparison_fusion}.

Our contributions can be summarized as follows:
\begin{enumerate}
    \item To the best of our knowledge, we propose the first streaming occupancy flow prediction framework that supports asynchronous multi-sensor fusion;
   
    \item  We design a novel temporal feature propagation strategy that achieves high performance on dense continuous occupancy prediction with only temporally sparse labels;
   
    \item Our proposed method achieves state-of-the-art performance on the widely used nuScenes~\cite{nuScenes} and Lyft L5~\cite{Lyft} datasets, validating the effectiveness and robustness of our proposed algorithms.
\end{enumerate}

% The remaining sections of this paper are structured as follows. Section.\ref{sec:related_works} presents a brief overview of related works. StreamingFlow method is elaborated in Section \ref{sec:methodology}. Experiments and results are analyzed and presented in Section \ref{sec:exps}. Section \ref{sec:conclu} concludes this paper.
\section{Related Works}\label{sec:related_works}
\begin{figure*}[t]
	\centering
	\includegraphics[width=0.9 \textwidth]{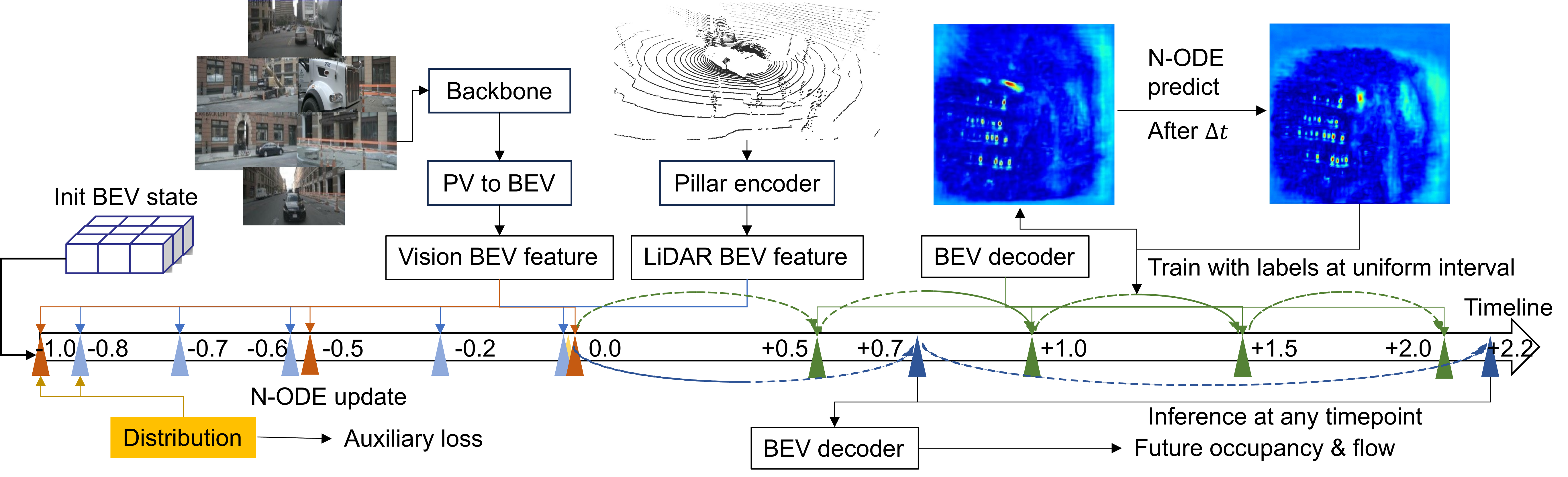}
	\caption{The framework of StreamingFlow. Raw data streams are encoded to BEV features, respectively. The SpatialGRU-ODE process operates on the timeline with two stages split by the present timestamp, asynchronous multi-sensor deep feature via the SpatialGRU-ODE update process and continuous occupancy flow prediction via SpatialGRU-ODE predict process.}
	\label{framework}
        \vspace{-1em}
\end{figure*}
\subsection{Occupancy forecasting}
Occupancy flow prediction, first proposed in~\cite{OccFlowField}, is a valuable supplement to trajectory-set prediction~\cite{FastandFurious,IntentNet}. The advantages of occupancy flow are a non-parametric distribution output and fine-grained grid flow. Recent LiDAR-based methods~\cite{MotionNet,PillarMotion,BE-STI} attach displacement vectors to non-empty BEV grids to describe short-term motion in future $1.0s$. These networks typically have a BEV-based spatial-temporal backbone and several shared heads that indicate the classification, motion, and displacement of each grid. STPN~\cite{MotionNet} and BE-STI~\cite{BE-STI} are effective BEV-based spatial-temporal feature extractors for binary-occupied grids which consist of only 2D convolutional blocks. Vision-centric pipelines usually formulate the prediction task as future instance segmentation on BEV grids. FIERY~\cite{FIERY} is the first vision-centric stochastic future prediction method. StretchBEV~\cite{StretchBEV} uses a variational autoencoder for decoupling learning of temporal dynamics and BEV decoder. End-to-end driving frameworks~\cite{ST-P3,BEVerse, UniAD, FusionAD} conduct planning based on occupancy prediction. UniAD~\cite{UniAD} and TBP-Former~\cite{TBP-Former} models temporal future with video transformers. One major bottleneck of video transformers is prediction of long-term sequence length~\cite{VideoTransformerSurvey}. Position encoding enables video transformer to interpolate shorter sequences but few video transformers can generalize to the unseen longer sequence length.  Unlike all prior works that predict discrete near-future snapshots, We focus on the streaming flow prediction which adapt the predictor to a more distant future and finer granularity.
\vspace{-0.375em}

\subsection{Multi-sensor Spatial-temporal Fusion}
Spatial multi-modal fusion is widely investigated in the feature level. BEVFusion and its variants~\cite{BEVFusion,FishingNet,RRF} combine multi-modal input features in a shared BEV space. \cite{UVTR,AutoAlignV2} support feature interaction in 3D voxel space. 
Temporal fusion is a common aggregator for perception tasks. LiDAR-based networks~\cite{BE-STI,MotionNet} typically employ data-level temporal fusion, which transforms multi-frame point clouds to the current ego's coordinate as inputs via ego pose or registration methods. Most vision-centric methods align the BEV feature maps from multiple time steps based on ego pose, and then concatenate them for fusion~\cite{BEVDet4D,BEVDepth,FIERY,SOLOFusion} or support their interactions of multi-frame BEV features through attention~\cite{BEVFormer,PolarFormer, UniFormer}. 
Prior spatial fusion methods usually assume strict synchronization. CoBEVFlow~\cite{CoBEVFlow} is most similar to our approach in that they predict BEV flow to interpolate asynchronous data timestamps from roadside unit to vehicle system, but short-term flow information cannot support long-term prediction. To this end, we provide a flexible spatio-temporal fusion framework which predicts the future frames and triggers update process as fusion when asynchronous sensor data arrives.

\section{Methodology}\label{sec:methodology}

% Inference of unknown states from derivatives estimation is a natural process with Euler, midpoint, or more complicated solvers provided by the TorchEq toolbox. A simple forward-Euler solver with initial value $h_{0}=h(x_{0})$ updates the state $y(t)$ iteratively,
% \begin{equation}
%     y(t+1)=y(t)+hf(h(t),t)
% \end{equation}

% The Midpoint solver is another alternative to ODE solvers, midpoint solver is calculated like,
% \begin{equation}
% y(t+1)=y(t)+hf\left[y(t)+\dfrac{h}{2}f(y(t),t),t+\dfrac{h}{2}\right]
% \end{equation}

\subsection{Framework}

% \begin{figure*}[ht]
% 	\centering
% 	\includegraphics[width=1 \textwidth]{fig/fusionmotion_framework.pdf}
% 	\caption{The framework of StreamingFlow. Raw data streams are first encoded to BEV features respectively. The SpatialGRU-ODE process operates on the timeline with two stages split by the present timestamp, asynchronous multi-sensor deep feature via the SpatialGRU-ODE update process and continuous occupancy flow prediction via SpatialGRU-ODE predict process.}
% 	\label{framework}
%         \vspace{-1em}
% \end{figure*}

The pipeline of StreamingFlow is depicted in \cref{framework}. Given multi-modal data streams from the previous few seconds, the task is defined as predicting the future several seconds' instance occupancy and flow on uniformly-shaped BEV grids. The framework includes three phases: 1) BEV encoders for LiDAR and camera branches, respectively, 2) asynchronous multi-sensor fusion, and 3) streaming occupancy prediction. As the data stream runs, modality-specific encoders map raw data to BEV features. The BEV state starts from zeroed features and fuses every incoming BEV feature in a trigger mode instead of a matching mode in conventional fusion. The BEV state propagates until a new BEV feature is received. The incoming BEV feature is fused to the BEV state and state continues to propagate until next incoming observation.

% The upper part is the feature extraction for LiDAR point clouds. We extract LiDAR features using a mixed representation of range view (RV) and bird's-eye view (BEV). Raw point clouds are densified from multi-frame sweeps and voxelized to binary 3D grids, after which they are passed through a spatial-temporal BEV-based CNN network to produce a BEV representation of non-empty pillars. Another path to raw points is mapping to the range view, using lightweight range-view networks for feature extraction, and projecting to BEV grids with a novel RV-Lift module for unseen or object parts in the rear. The two representations pass through a convolutional BEV fuser. The camera branch consists of a view projector from Lift,Splat\cite{LSS}, and a temporal aggregation block.

The SpatialGRU-ODE is the core for streaming prediction. The SpatialGRU-ODE iterates for ODE steps, updates when a new measurement occurs within a certain ODE step, and performs future prediction. In the training stage, the loss comes from the supervision of decoded BEV grids at the timestamps with labels, as well as an auxiliary probabilistic loss from Kullback-Leibler divergence (KLD) between updated BEV features and latent observations. In the inference process, SpatialGRU-ODE predicts via a variable ODE step only related to the required timestamps for evaluation and application.

The BEV decoder, loss functions, and post-processing techniques follow standard practices, and we leave the details in the supplementary materials.

\subsection{BEV Encoders}

\textbf{LiDAR branch.} We adopt the popular pillar representations for point cloud features. We observed that a very lightweight pillar-net encoder is sufficient for high accuracy.
Heavier backbones consisting of sparse convolutional modules or voxel transformers can be trivially integrated for performance at the cost of additional computation.

\textbf{Camera branch.} As a baseline for algorithms aligned with previous work, we adopt the Lift-splat-shoot (LSS~\cite{LSS}) settings with depth supervision for depth-based view transformation. We adopt accelerated BEV pool operators to speed up the inference stage. Other BEV mapping methods are trivially compatible with our framework.

\subsection{SpatialGRU-ODE}
Gate Recurrent Unit (GRU) consists of reset gate ($r_t$), update gate ($z_t$), and update vector ($g_t$). The update process follows the formula,
\begin{equation}
\small
  \begin{split}
  r_t&=\sigma(W_r x_t+U_r h_{t-1}+b_r) \\
  z_t&=\sigma(W_z x_t+U_z h_{t-1}+b_z) \\
  g_t&=tanh(W_h x_t+U_h (r_t \odot h_{t-1})+b_h) 
  \end{split}
  \label{gru_update}
\end{equation}

Neural-ODE on GRU blocks focuses N-ODE's ability to model continuous time series given sporadic sensor data coming from non-uniform time intervals. SpatialGRU-ODE is highly inspired by GRU-ODE-Bayes~\cite{gru_ode_bayes}, which serves as the foundation component of the proposed spatiotemporal fusion and flow predictor. \cref{gru_update} formulates the general process of GRU blocks. With regard to $h_t$ as a BEV feature, SpatialGRU has its state variable $h_t$ with shape $[B, C, H, W]$ where $B,C,H,W$ are the batch size, embedded dims, height, and width. $\sigma$ is the sigmoid function, and $tanh$ denotes a CNN block in SpatialGRU settings. $x_t$ is the input state at timestamp $t$ with the same shape as $h_t$. For simplicity, the update gate $W_r=U_r$, the reset gate $W_z=U_z$, and the GRU init bias $b_r = b_z$. $r_t \odot h_{t-1}$ is the matrix multiplication between reset gate and state. In summary, the update state $h_t$ is calculated as,
\begin{equation}
        \small
    h_t=z_t \cdot h_{t-1}+(1-z_t)\cdot g_t=GRU(h_{t-1},g_t)
\end{equation}
The amount of change of the state variable is calculated as:
\begin{equation}
        \small
\begin{array}{l}\Delta h_t=h_t-h_{t-1}=z_t\cdot h_{t-1}+(1-z_t)\cdot g_t-h_{t-1}\\ =(1-z_t)\cdot(g_t-h_{t-1})
\end{array}
\end{equation}
The derivatives of the state variable are calculated as:
\begin{equation}
        \small
    \frac{\Delta h(t)}{dt}=(1-z(t)) \cdot (g(t)-h(t))
    \label{gru_deri}
\end{equation}

GRU-ODE-Bayes has proven the convergence of this process, as is also applicable to the initial state of SpatialGRU-ODE. If $h_0\in[-1,1]$, then $h_j(t)\in [-1,1]$, where $j$ is the index of elements in $h(t)$, as $\left.\frac{d h(t)_{j}}{d t}\right|_{t:h(t)_{j}=1}\leq0$ and $\left.\dfrac{dh(t)_j}{dt}\right|_{t:h(t)_j=-1}\geq0$.

N-ODE has several nice properties when implemented on GRU models. The learned parametric derivative makes training not constrained by supervision from fixed future timestamps states, which greatly increases flexibility. On the other hand, unlike RNNs, which require emission intervals for updating, the GRU-ODE continuously defined dynamic model can naturally combine data observed at any given time.

The pseudo-code for SpatialGRU-ODE is shown in Algorithm 1. Given state-timestamp pairs $[t,m[t]]$ from previous multisensor features, The expected output is the states $[t,h[t]]$ of each measured past timestamp and anticipated future timestamp $t\in [t_0,...,t_{present},...,t_{future}]$. The ode step may be variable or constant. During the update procedure, a KL divergence loss $loss_{KLD}$ is gathered to quantify the similarity between updated state feature $h$ and observed feature $m[t_{obs}]$.
% \cref{sec:ode_update} and \cref{sec:ode_predict} introduce the two significant functions \textbf{ODE\_Update} and \textbf{ODE\_Predict}.
\begin{figure}[!t]
	\label{alg:spatialgru-ode}
 \small
	\renewcommand{\algorithmicrequire}{\textbf{Input:}}
	\renewcommand{\algorithmicensure}{\textbf{Output:}}
	% \removelatexerror
	\begin{algorithm}[H]
		\caption{The pseudo-code of SpatialGRU-ODE}
		\begin{algorithmic}[1]
			\REQUIRE state-timestamp pairs $[t,m[t]]$,$t\in [t_0,...,t_{present}]$, ODE step $\Delta t$          %%input
			\ENSURE $loss_{KLD}$, state-timestamp pairs $[t,h[t]]$,$t\in [t_0,...,t_{present},...,t_{future}]$.  %%output
			\STATE {Initialize current time $T=t_0$, KL divergence loss $loss_{KLD}=0$, current state feature $h=h_0$} 
            \FOR{$t_{obs}$ in $[t_0,...,t_{present}]$}
                \WHILE{$T<t_{obs}$}
                \STATE {$h$ = ODE\_Predict($h,\Delta t$)}
                \STATE {$T += \Delta t$}
                \ENDWHILE
                \STATE {$h,loss$ = ODE\_Update($h,m[t_{obs}]$)}
                \STATE {$loss_{KLD} += loss$}
                \STATE {save $h[t_{obs}] = h$}
            \ENDFOR
		    \FOR{$t_{predict}$ in $[t_{present+1},...,t_{future}]$}
                 \WHILE{$T<t_{predict}$}
                 \STATE {h = ODE\_Predict($h$,$\Delta t$)}
                 \STATE {$T += \Delta t$}
                 \ENDWHILE
                 \STATE {save $h[t_{predict}] = h$}
            \ENDFOR
		\end{algorithmic}
	\end{algorithm}
 \vspace{-1em}
\end{figure}

\subsection{Asynchronous Modality-agnostic Fusion}\label{sec:ode_update}

Asynchronous modality-agnostic fusion attempts to fuse LiDAR and camera BEV features in a way similar to the Beyasian predict-then-update process. The proposed strategy is able to eliminate two crucial limits for multi-sensor fusion, which are strict multi-modal synchronizations and fixed interval data flow.

% The two assumptions are often guaranteed by dataset creators during preprocessing, but they are not necessarily met in autonomous perception in the real world.

% \textbf{Cross-sensor synchronization.} On most datasets, synchronization is generally ensured by picking temporally nearby measurements as sensor data from the same sample. Synchronization is the essential underlying assumption of multi-sensor spatial alignment and feature fusion.

% \textbf{Uniform data flow.} The assumption that sensors operate consistently at all times is stringent. Uniform data flow in spatio-temporal networks will be disrupted by an excessively large burden on the vehicle's computing platform, communication congestion, and occasional sensor failure. Most fusion systems that rely on a single dominating sensor are susceptible to failure if that sensor occasionally malfunctions. Current fusion systems that handle different modalities equally, such as BEVFusion\cite{BEVFusion}, must switch models when switching input between LiDAR/camera/fusion mode, which is inflexible when one sensor experiences periodic errors and immediate recovery.

\begin{figure}[ht]
	\centering
	\includegraphics[width=0.45\textwidth]{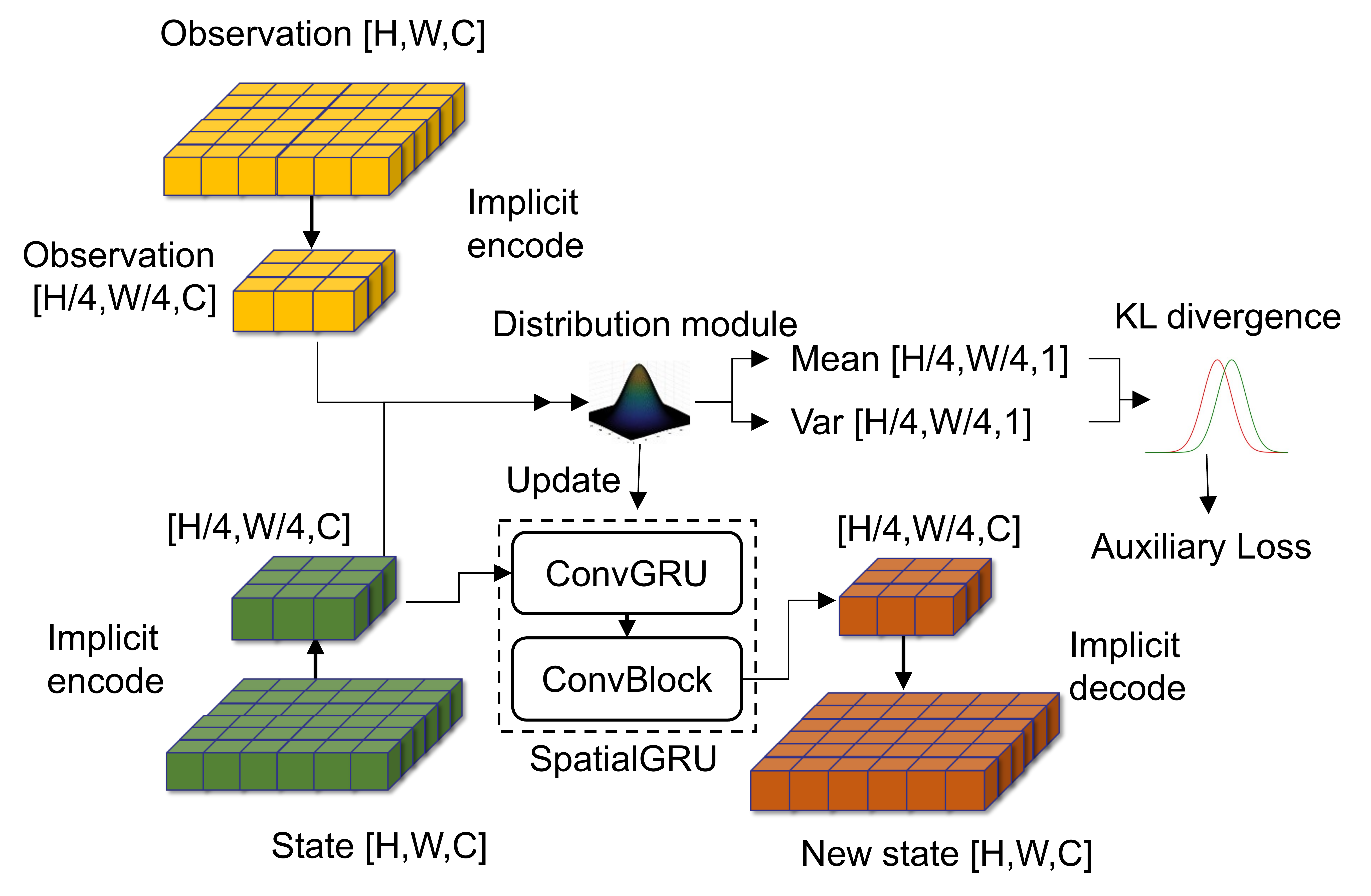}
	\caption{Illustration of the measurement update process of SpatialGRU-ODE in temporal-agnostic fusion.}
	\label{ode_update}
\end{figure}

\cref{ode_update} presents the network structure for the SpatialGRU-ODE update process. After raw data is encoded to BEV maps by different BEV encoders, BEV features from different modalities are agnostic on the source, ordered chronologically, and formulated as multiple timestamp-BEV-feature pairs $[t,h_t]$. The output retains the same shape as the input, but the hidden features are as $\frac{1}{4}$ size of the input for the sake of eliminating memory storage. In order not to degrade the ultimate performance, both state and observation in BEV feature space with shape $[B, C, H, W]$ are implicitly encoded into a smaller feature space $[B,C,H/4,W/4]$ and finally implicit decoded in the same way for occupancy flow decoder. The temporal propagator runs the subsequent ODE step till the arrival of the next observation. When a new observation occurs, the predicted state and new observation are processed into the distribution module with CNN blocks to obtain the mean and variance of BEV features both with the size $[1, H/4, W/4]$. Means and variances of both features are estimated using KL divergence, an auxiliary loss that reflects the similarity between predicted and measured characteristics. Probabilistic auxiliary loss is calculated as follows:
\begin{equation}
  loss_{KLD} = \frac{1}{H\times W} \sum_{j=1}^{H\times W}D_{K L}(p_{\mathrm{state},j}||p_{\mathrm{meas},j})
\end{equation}

The core module for the updated BEV feature is a dual-pathway SpatialGRU module, which first propagates the observed and predicted features respectively, then mixes the two hidden states distributions and finally undergoes the weighted summation of two features after the trusting gate and softmax function to generate the final new state.

\subsection{Streaming Occupancy Flow Predictor}\label{sec:ode_predict}

The occupancy flow predictor handles multi-frame BEV features from past and present timestamp-agnostic fusion as input and propagates to future steps in an ODE-enabled variational recurrent neural network. Particularly, the granularity of temporal prediction is decoupled from the training granularity and only related to the minimal ODE step in this method. If we set the variable ODE step, the prediction enables arbitrary granularity. 

% Prior arts with RNNs are hard-coded for emission intervals in future prediction, and at the same time, labels of these time steps are generated for supervision. However, the proposed continuous predictor has the advantage of predicting future occupancy for any given time step even if some time steps are not provided with label supervision. In essence, It models the unit-time-changing trend of BEV occupancy flow.

\begin{figure}[ht]
	\centering
	\includegraphics[width=0.45\textwidth]{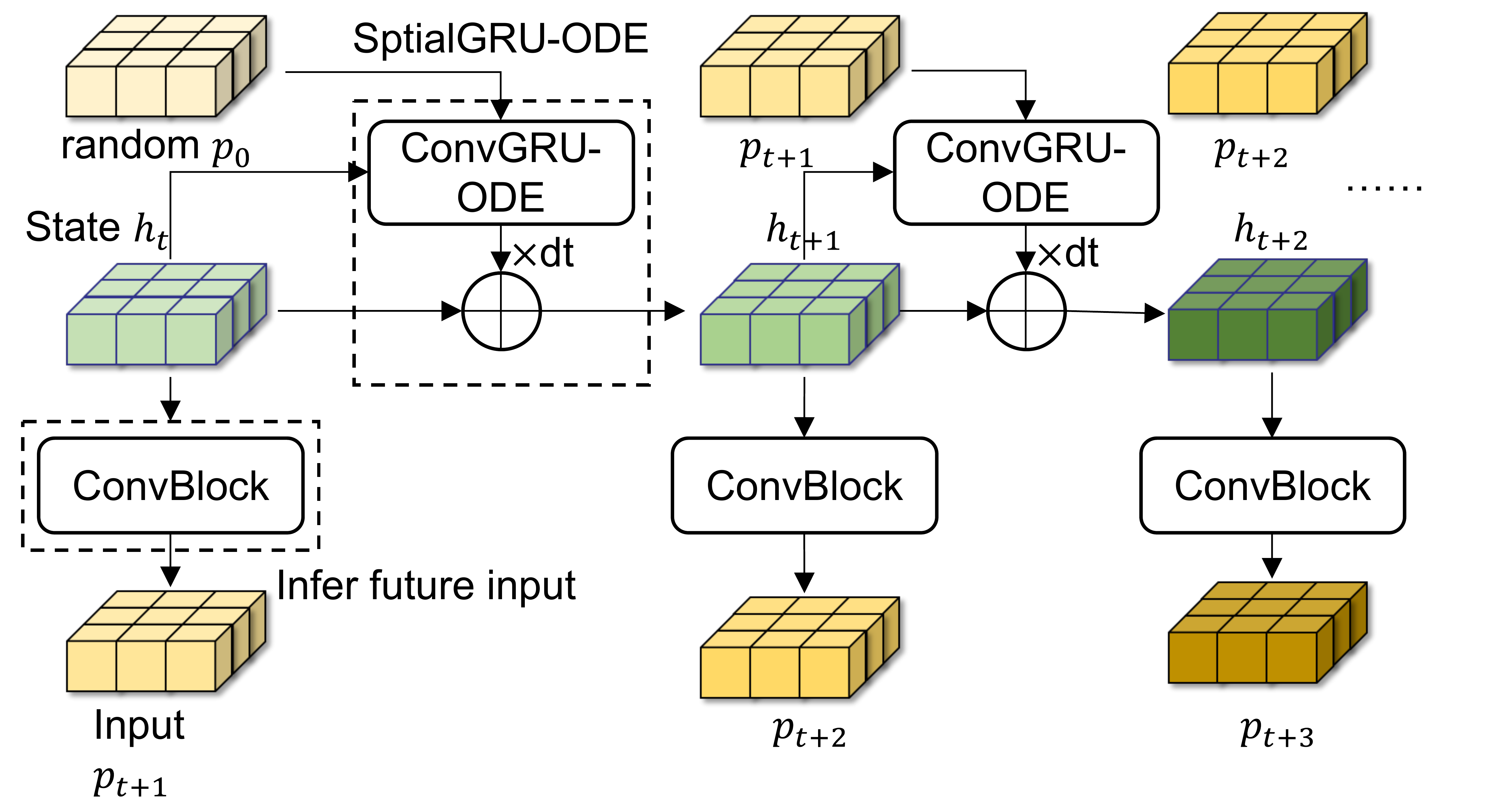}
	\caption{Illustration of streaming prediction process of SpatialGRU-ODE.}
	\label{ode_predict}
\end{figure}

Fig. \ref{ode_predict} illustrates the SpatialGRU-ODE procedure. Similar to the update process, the states are implicitly encoded into a smaller feature space. Except for the initial input which is the first BEV feature, the input with shape $[H/4,W/4,C]$ is inherited from the last ODE step. The core module, SpatialGRU-ODE, computes the derivatives of BEV features in unit time intervals via a recurrent block specified in \cref{gru_deri}. SpatialGRU-ODE may employ either Euler or Midpoint solver for the update and fixed or variable ODE time step. To obtain the projected states, the derivatives are multiplied by delta time $dt$ and added to the initial states. A CNN block is used to infer the next input from the current state in order to make the subsequent prediction. Only states near to the needed timestamps with supervision (closer than half of the minimum ode step) are preserved and decoded in the BEV decoder for supervision. Thus, the prediction time step $dt$ is not always constant and uniform, and supervision signals are not required for each future timestamp.

\section{Experiments}\label{sec:exps}

In this section, we seek to answer two questions with the following experiments: (1) Whether the model works as well as published state-of-the-art algorithms for standard occupancy forecasting tasks. (2) Whether we are able to generalize to streaming forecasting settings with one model trained with temporally-sparse occupancy labels.

\begin{figure*}
%\begin{figure*}[ht]
	\centering
	\includegraphics[width=1.0 \textwidth]{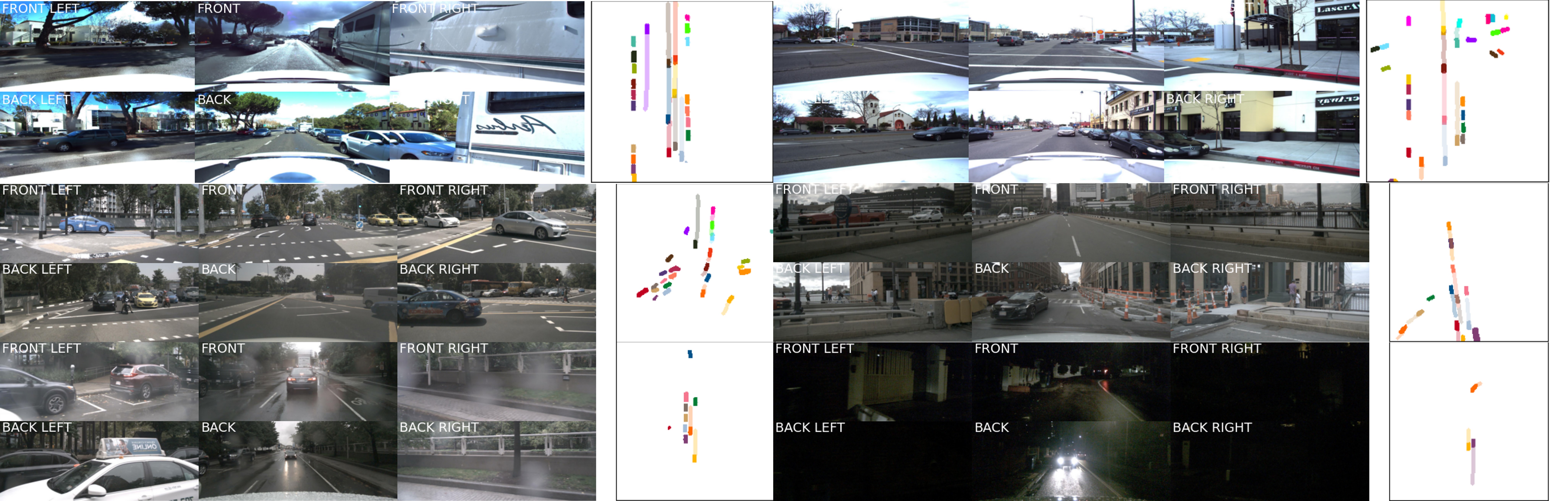}
	\caption{Visualization of StreamingFlow for diverse driving scenarios. Different colors represent different instances of the agents, and lighter colors represent the future occupancy of the agents. (top): samples from Lyft dataset, highway (left), and urban (right). (middle and bottom): samples from nuScenes dataset, sunny (middle left), overcast (middle right), rainy (bottom left), and night (bottom right). StreamingFlow works well in all challenging driving scenarios.}
	\label{visualization_examples}
\end{figure*}

\subsection{Datasets and Metrics}

The NuScenes~\cite{nuScenes} dataset is a public large-scale autonomous driving dataset collected by Motional. NuScenes dataset provides a full sensor suite including 1 top LiDAR, 6 cameras, 5 radars, GPS, and IMU, with 360° coverage of the surroundings. It contains 1000 scenes, each lasting 20 seconds, of which annotations of 850 scenes at 2Hz are available. We use the training and validation splits as same as previous works~\cite{BE-STI,FIERY,BEVerse,ST-P3,StretchBEV}, with 28130 samples for training and 6019 samples for validation.

The Lyft L5 dataset~\cite{Lyft} is another large-scale autonomous dataset provided by Lyft. This dataset contains 180 scenes, each lasting 25-45s in length and annotated at 5Hz. We use 6 ring cameras and the top LiDAR from its sensor suite, which has a 360° field of view. We use training and validation splits from the practice of FIERY~\cite{FIERY}, which include 16000 training samples and 4000 validation samples.

The occupancy flow prediction is formulated as a video panoptic prediction task~\cite{VPQ}. We use Intersection over Union (IoU) to measure the semantic segmentation quality of each frame, and video panoptic quality (PQ), recognition quality (RQ), and segmentation quality (SQ) to measure both the accuracy and consistency of instance segmentation. VPQ is computed as:

\vspace{-1em}

\begin{equation}
    \text{VPQ}=\sum_{t=0}^H\dfrac{\sum_{(p_t,q_t)\cdot TP_t}\text{IoU}(p_t\cdot q_t)}{|TP_t|+\dfrac{1}{2}|FP_t|+\dfrac{1}{2}|FN_t|}
    % \small
    % \begin{array}{c}\text{VPQ}=\sum_{t=0}^H\dfrac{\sum_{(p_t,q_t)\cdot TP_t}\text{IoU}(p_t\cdot q_t)}{|TP_t|+\dfrac{1}{2}|FP_t|+\dfrac{1}{2}|FN_t|}\\ =\sum_{t=0}^H\dfrac{|TP_t|}{|TP_t|+\dfrac{1}{2}|FP_t|+\dfrac{1}{2}|FN_t|}=\text{SQ}\cdot \text{RQ}\end{array}
\end{equation}

\subsection{Implementation Details}
\textbf{Task details.}
The standard occupancy prediction task is to take the past 1.0s sensor data for input, which is 3 keyframes in nuScenes (5 keyframes in Lyft), predict the motion in the future 2s, which is 4 keyframes in nuScenes and 10 keyframes in Lyft. 

\textbf{Data details.} The BEV occupancy labels are from 3D bounding boxes labels projected to BEV. The perception range is set as $[100m,100m]$ and the grid resolution is set as $0.5m$. Images are resized to $224\times480$ pixels for each frame. Each point cloud is densified by aggregating point clouds of adjacent sweeps. Point clouds are voxelized to $200 \times 200 \times 13$ voxels. StreamingFlow adopts an asynchronous data stream as input, with LiDAR streams every 0.2s and camera streams every 0.5s as input. For larger models fairly compared with the state-of-the-art, we use images with size $320\times800$ and aggregated LiDAR streams, each with 10 sweeps. 

\textbf{Model details.}
The framework is implemented on Pytorch 1.10.2 and Pytorch Lightning 1.2.5. The optimizer is AdamW with a learning rate of $1e-4$. The learning strategy is cosine annealing with a weight decay of $0.01$. All models are trained on 8 A6000 GPUs for 20 epochs with a total batch size 16. The image backbone is EfficientNet-b4~\cite{efficientnet} for tiny and base model and Effi-b7 for small model.

\begin{table*}[htbp]
	\centering
        \small
	\begin{tabular*}{\textwidth}{@{\extracolsep{\fill}}l|c|cc|c|ccc}
		\toprule
		Method & Modality & Image backbone & LiDAR backbone & Future Semantic  Seg.  & \multicolumn{3}{c}{Future Instance Seg.} \\
		& & & & IoU↑  & PQ↑   & SQ↑   & RQ↑ \\
		\midrule
		Static* & C & Effi-B4 & - & 32.2  & 27.6  & 70.1  & 39.1 \\
		FIERY*\cite{FIERY} & C & Effi-B4 & -  & 37.0  & 30.2  & 70.2  & 42.9 \\
		StretchBEV\cite{StretchBEV} & C & Effi-B4 & - & 37.1  & 29.0  & -  & - \\
		BEVerse\cite{BEVerse} & C & Swin-tiny & - & 38.7  & 33.3  & 70.6  & 47.2 \\
            BEVerse\cite{BEVerse} & C & Swin-small & - & 40.8  & 36.1  & 70.7  & 51.1 \\
		ST-P3 Gaus.\cite{ST-P3} & C & Effi-B4 & - & 38.6  & 31.7  & 70.2  & 45.2 \\
		ST-P3 Ber.\cite{ST-P3} & C & Effi-B4 & - & 38.9  & 32.1  & 70.4  & 45.6 \\
            UniAD\cite{UniAD} & C & ResNet-101 & - & 40.2  & 33.5  & -  & - \\
		\midrule
            MotionNet\cite{MotionNet} & L & - & STPN & 37.2  & 38.9  & 75.6  & 51.4 \\ 
            BE-STI\cite{BE-STI} & L & - & BESTI-STPN & 40.1  & 41.0 & 75.5  & 54.3 \\
              \midrule
		% StreamingFlow & Effi-B4 + RV-Lift & 43.0  & 36.3  & 71.2  & 51.5 \\
  % 		StreamingFlow & Effi-B4 + STPN & 48.7  & 45.8  & 74.6  & 61.3 \\
  %     	StreamingFlow-ODE & Effi-B4 + STPN & 47.8  & 42.9  & 73.5  & 58.6 \\

           StreamingFlow-tiny & LC & Effi-B4 & PillarNet & 47.8  & 46.1  & 75.8  & 60.8 \\
  		StreamingFlow-small & LC & Effi-B7 & PillarNet  & 51.1  & 48.0  & 75.1  & 64.0 \\
    
    	FusionAD\cite{FusionAD} & LC & ResNet-101 & VoxelNet & 51.5  & 51.1  & -  & - \\    
          StreamingFlow-base & LC & Effi-B4 & VoxelNet  & 53.9  & 52.8  & 76.8  & 68.7 \\

		\bottomrule
	\end{tabular*}%
 	\caption{Comparison with the state-of-the-art methods for instance-aware occupancy flow prediction on nuScenes\cite{nuScenes} validation set. 'C' denotes vision-only methods, 'L' denotes LiDAR-based methods, and 'LC' denotes a LiDAR-camera fusion methods. $*$: reported in \cite{FIERY}.}
	\label{tab:comparison_nusc}%
\end{table*}

\subsection{Comparison with the State-of-the-art}
Comparisons of methods implemented in the nuScenes dataset are shown in \cref{tab:comparison_nusc}. All methods are trained for a standard instance flow segmentation and prediction task. \cref{visualization_examples} illustrates some examples covering diverse scenarios.

StreamingFlow achieves competitive performance while significantly outperforming other single-modality methods. StreamingFlow surpasses the best vision-based BEVerse~\cite{BEVerse} with a larger backbone Swin-small~\cite{Swin} by $+7.0$ IoU and $+10.0$ VPQ, and the best LiDAR-based BE-STI~\cite{BE-STI} $+7.7$ IoU and $5.1$VPQ. Compared to FusionAD, StreamingFlow-base uses the same LiDAR backbone(SpConv), similar image backbone(Effi-B4 vs. ResNet101) but a half smaller image size($800\times{}320$ vs. $1600\times{}900$), and is $+1.7$ VPQ and $2.4$ IoU higher than the latest fusion-based FusionAD~\cite{FusionAD}.

Comparisons of methods implemented in the Lyft L5 dataset are shown in \cref{tab:comparison_lyft}. StreamingFlow achieves the best of all metrics while surpassing baseline methods by a large margin: $+20.6$ IoU and $+23.5$ VPQ against ST-P3~\cite{ST-P3}. Moreover, it surpasses our reproduced BEVFusion with lightweight encoders by $+2.3$ mIoU and $0.2$ VPQ.

\subsection{Streaming Forecasting Results} 

We design three experiments on nuScenes dataset for demonstration of the streaming occupancy prediction: (1) Extension of the prediction temporal horizons to unseen future; (2) Prediction at any desired future time point; (3) Data streams input at different frame rates. Experiments are conducted on models loaded from the checkpoints trained on standard instance forecasting tasks.
\begin{table}[tbp]
	\centering
        \small
	\setlength{\tabcolsep}{9pt} % 调整列距
	\begin{tabular}{l|c|ccc}
		\toprule
		Method     &      IoU↑  & PQ↑   & SQ↑   & RQ↑ \\
	
		\midrule
		Static*  & 24.1  & 20.7  & -  & - \\
            Extrapolation model*  & 24.8  & 21.2  & -  & - \\ 
		FIERY*\cite{FIERY}  & 36.3  & 27.2 & -  & - \\
		ST-P3\cite{ST-P3}  & 36.3  & 32.4 & 71.1  & 45.5 \\
            \midrule
		% StreamingFlow & Effi-B4 + STPN & 54.6  & 55.7  & 78.0  & 71.4 \\
  		BEVFusion-tiny$\dagger$  & 54.6  & 55.7  & 78.0  & 71.4 \\
  		StreamingFlow-tiny  & 56.9  & 55.9  & 78.1  & 71.6 \\
		\bottomrule
	\end{tabular}%
 	\caption{Comparison with the state-of-the-art methods for future instance segmentation (2.0s) on Lyft L5 AV~\cite{Lyft} validation set. $*$: results reported in~\cite{FIERY}. $\dagger$: results of BEVFusion with LSS-based image encoder and pillar-based encoder reproduced by us.}
	\label{tab:comparison_lyft}%
 \vspace{-1em}
\end{table}

\textbf{Extension of the temporal horizons of unseen future.} \cref{tab:longer_horizon} shows the prediction result of StreamingFlow which extends the temporal horizons from 2.0s to 8.0s. Baselines reported in \cite{StretchBEV} are trained for different lengths of temporal horizons. StreamingFlow shows excellent performance in variable prediction length without retraining. Similar to prior arts which extend prediction horizons to farther future, StreamingFlow shows a similar gradual performance decay. We then compare zero-shot Streamingflow with the model trained with the longest future 8s groundtruth labels in \cref{tab:comparison_zeroshot}. The zero-shot long-term prediction results are only $-0.9$ IoU and $-1.8$ VPQ less than fully supervised models, which demonstrates great generalization ability of foreseeing the unseen future.

\begin{table}[htbp]
	\centering
        \small
	\setlength{\tabcolsep}{7.5pt} % 调整列距
	\begin{tabular}{l|c|cccc}
		\toprule
		% \multirow{2}[2]{*}{Method} & Backbone & Future Semantic  Seg.  & \multicolumn{3}{c}{Future Instance Seg.} \\
		% & & IoU↑  & PQ↑   & SQ↑   & RQ↑ \\
           Method    &  Pred  &    IoU↑  & PQ↑   & SQ↑   & RQ↑ \\
           \midrule
            \multirow{3}{*}{FIERY}        & 2s  & 35.8  & 29.0 & -  & - \\  
                   & 4s  & 30.1  & 23.6 & -  & - \\
                   & 6s  & 26.7  & 20.9 & -  & - \\
           \midrule
            \multirow{3}{*}{StretchBEV}        & 2s  & 37.1  & 29.0 & -  & - \\  
                    & 4s  & 32.5  & 23.8 & -  & - \\
                    & 6s  & 28.4  & 21.0 & -  & - \\
		\midrule
            \multirow{5}{60pt}{StreamingFlow}       & 2s  & 47.8  & 46.1 & 75.8  & 60.8 \\  
            & 3s  & 44.4  & 41.1 & 74.1  & 55.4 \\
            & 4s  & 41.7  & 38.4 & 73.6  & 52.1 \\
            & 5s  & 39.1  & 35.8 & 73.2  & 48.9 \\
            & 6s  & 36.7  & 33.6 & 72.9  & 46.0 \\
            & 8s  & 32.5  & 29.8 & 72.6  & 41.0 \\
		\bottomrule
	\end{tabular}%
 	\caption{Comparison of occupancy flow prediction for any temporal horizons.}
	\label{tab:longer_horizon}%
 \vspace{-1em}
\end{table}

\begin{table}[htbp]
	\centering
        \small
	\setlength{\tabcolsep}{3pt} % 调整列距
	\begin{tabular}{l|ccccc}
		\toprule
		Schedule & 2s & 3s & 4s & 6s  & 8s  \\
		\midrule
          ZS. & 47.8/46.1 & 44.4/41.1 & 41.7/38.4 & 36.7/33.6 & 32.5/29.8 \\
          Sup. & 48.2/46.9 & 46.0/45.2 & 41.9/38.7 & 37.8/34.2 & 33.6/31.6   \\
  		% StreamingFlow-base & LC & Effi-B7 (1600x640) & PillarNet  &  \\
		\bottomrule
	\end{tabular}%
        \vspace{-1em}
 	\caption{ZS: zero-shot inference of StreamingFlow. Sup: StreamingFlow trained with the longest prediction horizons 8s. Each value is given in the format of `IoU/PQ'.}
	\label{tab:comparison_zeroshot}%
\end{table}

\textbf{Streaming prediction at any desired time point.}
Since nuScenes do not provide labels for non-keyframes, we adopt the practice of MotionNet and interpolate the box instances between two adjacent keyframes. We randomly choose to predict at a list of time points and report the results in \cref{tab:stream_timepoint}. To simply illustrate the capability, we designed the main task to predict uniformly spaced temporal horizons, but this framework can predict uneven and arbitrary future snapshots. The prediction results are stable generally. There is a slight decline as the forecast becomes denser.  

\begin{table}[htbp]
	\centering
        \small
	\setlength{\tabcolsep}{10pt} % 调整列距
	\begin{tabular}{l|c|ccc}
		\toprule
		% \multirow{2}[2]{*}{Method} & Backbone & Future Semantic  Seg.  & \multicolumn{3}{c}{Future Instance Seg.} \\
		% & & IoU↑  & PQ↑   & SQ↑   & RQ↑ \\
           Prediction Interval     &      IoU↑  & PQ↑   & SQ↑   & RQ↑ \\
           \toprule
		
            0.5  & 47.8  & 46.1  & 75.8  & 60.8 \\
            
		0.25  & 43.4  & 40.1  & 74.1  & 54.1 \\

            0.6  & 45.6  & 44.3  & 75.2  & 58.9 \\
            
            % Custom & 41.2  & 38.9  & 73.9  & 52.6 \\
		\bottomrule
	\end{tabular}%
 	\caption{Prediction results at any desired intervals. For example, prediction at the interval of 0.6 means prediction at $[0.6, 1.2, 1.8]$.}
	\label{tab:stream_timepoint}%
 \vspace{-1em}
\end{table}

 We provide demo videos as supplementary materials to show the effectiveness of streaming prediction. We provide a finer granularity of streaming perception at any future time. In the video, we visualize the prediction results every 0.05s from the current to the future 2s to 3s, with a total of 40 to 60 frames.

\textbf{Asynchronous multi-modal data streams as input.} We validate asynchronous data streams as input and standard occupancy forecasting as output. We test multiple settings of data stream frame rate and report the results in Tab. \ref{tab:stream_frame_rate}. The model performs the best when the inference uses the training configuration. Denser inputs lead to similar forecasting accuracy.

\begin{table}[htbp]
	\centering
        \small
	\begin{tabular}{lc|c|ccc}
		\toprule
		% \multirow{2}[2]{*}{Method} & Backbone & Future Semantic  Seg.  & \multicolumn{3}{c}{Future Instance Seg.} \\
		% & & IoU↑  & PQ↑   & SQ↑   & RQ↑ \\
           LiDAR stream     &  Cam stream  &    IoU↑  & PQ↑   & SQ↑   & RQ↑ \\
           \toprule

            5 & 2  & 47.8  & 46.1  & 75.8  & 60.8 \\
            
		  10 & 2  & 47.6  & 45.8  & 75.6  & 60.5 \\

            10 & 4  & 47.1  & 45.4  & 75.1  & 60.4 \\
            
		\bottomrule
	\end{tabular}%
 	\caption{Prediction results with multi-modal data stream of different frame rates. Frame rates are shown in Hz.}
	\label{tab:stream_frame_rate}%
\end{table}

\subsection{Ablation Study}

\textbf{Effect of different fusion strategies.} We compare SpatialGRU-ODE with fusion strategies at different stages and the implementation details are in supplementary material. As shown in \cref{tab:compare_fusion}, asynchronous fusion is inferior to synchronous fusion and SpatialGRU-ODE for both datasets. For nuScenes dataset, the ordinary temporally-synchronized fusion approach performs the best. SpatialGRU-ODE performs $-2.4$ IoU and $-0.9$ VPQ, slightly less than BEVFusion~\cite{BEVFusion}. For Lyft dataset, SpatialGRU-ODE outperforms synchronized fusion. It performs $+2.3$ IoU and $+0.2$ VPQ more than BEVFusion~\cite{BEVFusion}.

\begin{table}[htbp]
	\centering
        \small
	\setlength{\tabcolsep}{9pt} % 调整列距
	\begin{tabular}{l|c|ccc}
		\toprule

           Fusion mode     &      IoU↑  & PQ↑   & SQ↑   & RQ↑ \\
           \midrule       
        \multicolumn{4}{c}{nuScenes dataset} \\
		\midrule
		Spatial then temporal & 50.2  & 47.0  & 75.6  & 62.2 \\
            Temporal then spatial  & 44.7  & 42.7  & 74.6  & 57.2 \\ 
            SpatialGRU-ODE & 47.8  & 46.1  & 75.8  & 60.8\\
		\midrule
        \multicolumn{4}{c}{Lyft L5 AV dataset} \\
		\midrule
		Spatial then temporal & 54.6  & 55.7  & 78.0  & 71.4 \\
            Temporal then spatial  & 50.4  & 50.2  & 75.2  & 66.7 \\ 
            SpatialGRU-ODE  & 56.9  & 55.9  & 78.1  & 71.9 \\
		\bottomrule
	\end{tabular}%
 	\caption{Comparison with the baseline fusion methods for future instance segmentation (2.0s) on  nuScenes validation set. StreamingFlow-ODE is compared with different fusion modes. Spatial fusion is conducted in the same way as BEVFusion~\cite{BEVFusion}. `Spatial Then Temporal' is for synchronous mode, and `Temporal Then Spatial' is for asynchronous mode.}
	\label{tab:compare_fusion}%
 \vspace{-1em}
\end{table}

\textbf{Effect of ODE solvers.} ODE solvers are vital tools for a N-ODE framework. We test two basic ode solvers, Euler, and midpoint solver as compared in \cref{tab:compare_solver} on two datasets. In both cases, the midpoint solver yields superior VPQ, surpassing the Euler solver by $3.2$ points on nuScenes dataset and $0.8$ points on Lyft dataset. On nuScenes dataset, the midpoint solver shows similar IoU with the Euler solver, but on Lyft dataset, the midpoint solver performs worse semantic segmentation by $-1.9$. 

\begin{table}[htbp]
	\centering
        \small
	\setlength{\tabcolsep}{12pt} % 调整列距
	\begin{tabular}{l|c|ccc}
		\toprule
		% \multirow{2}[2]{*}{Method} & Backbone & Future Semantic  Seg.  & \multicolumn{3}{c}{Future Instance Seg.} \\
		% & & IoU↑  & PQ↑   & SQ↑   & RQ↑ \\
           ODE Solver     &      IoU↑  & PQ↑   & SQ↑   & RQ↑ \\
           \midrule       
        \multicolumn{4}{c}{nuScenes dataset} \\
		\midrule
            Euler  & 47.8  & 42.9  & 73.5  & 58.6 \\
            Midpoint  & 47.8  & 46.1  & 75.8  & 60.8 \\
		\midrule
        \multicolumn{4}{c}{Lyft L5 AV dataset} \\
		\midrule
            Euler  & 56.9  & 55.9  & 78.1  & 71.5 \\
            Midpoint  & 55.0  & 56.7  & 78.2  & 72.4 \\
		\bottomrule
	\end{tabular}%
 	\caption{Comparison with prediction using different ODE solvers.}
	\label{tab:compare_solver}%
 \vspace{-1em}
\end{table}

\textbf{Effect of ODE update intervals.} We ablate different ODE update intervals which determine the minimum granularity of occupancy prediction. The ODE step is set to $0.05$, $0.1$, $0.5$ and variable for comparison. Variable means the ODE step is the next timestamp minus the current timestamp. In general, the finer a single update is divided during the ODE update process, the higher the final prediction accuracy. It is also intuitive that a higher update frequency may eliminate continuous errors. As shown in \cref{tab:continuous_prediction}, the predictor with interval $0.05s$ surpasses interval $0.1s$ by $0.1$ IoU and $3.1$ VPQ, and interval $0.5s$ by $0.5$ IoU and $3.1$ VPQ. Remarkably, the variable time step update also shows nice performance, which is $+0.5$ IoU and $-1.9$ VPQ when compared to the predictor with ODE step $0.05s$. A variable time step updater strikes a good balance between precision and inference speed, so it is more favorable for long-term prediction to 8s future.

\begin{table}[htbp]
	\centering
        \small
	\setlength{\tabcolsep}{13pt} % 调整列距
	\begin{tabular}{l|c|ccc}
		\toprule
		% \multirow{2}[2]{*}{Method} & Backbone & Future Semantic  Seg.  & \multicolumn{3}{c}{Future Instance Seg.} \\
		% & & IoU↑  & PQ↑   & SQ↑   & RQ↑ \\
           Step     &      IoU↑  & PQ↑   & SQ↑   & RQ↑ \\
           \toprule
           \multicolumn{4}{c}{nuScenes dataset} \\
		\midrule
		0.05  & 47.8  & 46.1  & 75.8  & 60.8 \\
            0.1  & 47.7  & 43.0  & 73.7  & 58.4 \\
            0.5  & 47.3  & 42.1  & 73.3  & 57.4 \\
            Variable & 48.2  & 44.2  & 73.9  & 59.8 \\
            \midrule
            \multicolumn{4}{c}{Lyft L5 dataset} \\
            \midrule
             0.05  & 56.9  & 55.9  & 78.1  & 71.6 \\
             Variable & 54.0  & 53.5  & 77.1  & 69.7 \\
		\bottomrule
	\end{tabular}%
 	\caption{Comparison of prediction under different ode steps.}
	\label{tab:continuous_prediction}%
 \vspace{-2em}
\end{table}

% The difference from comparison with BEVFusion on different datasets is possibly related to the sparsity of supervision. On nuScenes dataset, the predictor is supervised approximately every $0.5s$, while the internal ode step is $0.05s$, which means the predictors need to update 10 rounds before strong supervision. On Lyft dataset, the predictor is supervised every 0.2s, which only needs 4 rounds for possible supervision. Though fully-dense supervision at each time step is unaffordable, sparser supervision may impair the final prediction performance. 
\section{Conclusion}\label{sec:conclu}
We present StreamingFlow as the first practice for fusion-based streaming occupancy forecasting. Starting from the motivation of fusing and predicting from non-ideal data streams, the streaming feature brings a more flexible temporal scene understanding and generalizes occupancy forecasting well to any temporal horizon. The advantage of SpatialGRU-ODE is that it decouples the supervision in the training process and outputs in the inference process, by modeling the derivatives of BEV grids and propagating future grid states to required timestamps. We hope that this work will inspire more Neural-ODE applications and insights into the streaming automotive perception.
\section{Acknowledgement}
This work was supported in part by the National Natural Science Foundation of China under Grants (52372414, U22A20104, 52102464). This work was also sponsored by Tsinghua University-DiDi Joint Research Center for Future Mobility and Tsinghua University-Zongmu Technology Joint Research Center.

{
    \small  \bibliographystyle{ieeenat_fullname}
    \bibliography{main}
}

% WARNING: do not forget to delete the supplementary pages from your submission 
\clearpage
\setcounter{page}{1}
\maketitlesupplementary

\section{More model designs}
\subsection{Preliminary}
\textbf{Neural-ODE fundamentals.} The Neural Ordinary Differential Equation (N-ODE) is proposed for learning some simple continuous implicit processes in dynamic systems. The core characteristic is that it uses neural networks to parameterize derivatives of hidden states $f(h(t),t,\theta)$ instead of specifying a discrete sequence of hidden layers $h(t)$. Given a continuous process $h_t$, the update process is to calculate the derivatives $\frac{d h(t)}{dt}$ via a neural network,
\begin{align}
h(t+1)&=h_t+f(h_t,\theta_t)\\
\frac{d h(t)}{dt}&=f(h(t),t,\theta)
\end{align}

\subsection{Decoders, Loss and Post-processing}
The decoders are inherited from FIERY\cite{FIERY,StretchBEV,ST-P3}. Five distinct decoders produce centerness regression, BEV segmentation, offset to the centers, future flow vectors, and instance given the BEV feature representation of past and future frames. The shared BEV feature will be input into a shared Resnet18 BEV backbone and five independent CNN blocks.

The loss design consists of spatial regression loss, segmentation loss, and probabilistic loss. Spatial regression loss is responsible for regressing centerness and offsets in a L1 loss or mean square error(MSE/L2) loss manner. Segmentation loss is the computation of the cross-entropy loss on multi-frame BEV semantic grids from the past to the future. Probabilistic loss computes the divergence between updated BEV features and measurement features with regard to their mean and variance on BEV grids. The overall loss is calculated as follows:
\begin{equation}
    Loss = \lambda_1 * L_{seg} + \lambda_2 * L_{spatial} + \lambda_3 * L_{kld}
    \label{loss}
\end{equation}
In \cref{loss}, $\lambda$ is the weight of each loss. We use uncertainty to balance different losses at the training stage.

\section{More Experiments}
\subsection{Runtime Analysis}
Table. \ref{table_runtime_analysis} compares the run-time training memory of the proposed method with baselines in different dataset settings. As the time interval becomes denser, the training cost for standard GRU units becomes unaffordable, whereas the proposed method is more adaptable. For standard 4-keyframe supervision, StreamingFlow only requires $+2G$ more memory compared to a BEVFusion-style implementation. As the density of supervision signals increases, the ODE approach requires $-4G$ less memory than standard GRU counterparts.

The inference speed of StreamingFlow is measured by the average time required to process validation samples over 250 forward passes on a laptop equipped with a single RTX3090. As StreamingFlow inherits the same framework and modules from FIERY\cite{FIERY}, it is compared with FIERY\cite{FIERY} and StretchBEV\cite{StretchBEV}. For the settings of the standard task and variable ode steps, StreamingFlow runs at 0.1968s/sample, faster than FIERY(0.6436s/sample) and StretchBEV(0.6469s/sample) reported in \cite{StretchBEV}. The SpatialGRU-ODE works at a similar speed with prior temporal modules. The inference speed for tasks with finer granularity (40-frame experiment) is around 0.5s per sample. Obviously, higher prediction frequencies harm the run-time delay. Therefore, sparse streaming prediction based on variable ODE step by request is recommended.

\begin{table}[htbp]
	\centering
	\setlength{\tabcolsep}{5pt} % 调整列距
	\begin{tabular}{lcc|c}
		\toprule
		% \multirow{2}[2]{*}{Method} & Backbone & Future Semantic  Seg.  & \multicolumn{3}{c}{Future Instance Seg.} \\
		% & & IoU↑  & PQ↑   & SQ↑   & RQ↑ \\
           Dataset     &      Config &      Supervised frames & Memory   \\
		\midrule
		nuScenes  & GRU-base  & 4    & 11G \\
		nuScenes  & GRU-ODE  & 4    & 13G \\   
  		nuScenes  & GRU-base  & 40    & 39G \\
		nuScenes  & GRU-ODE  & 40    & OOM \\ 
		\midrule
		Lyft  & GRU-base  & 10   & 28G \\
		Lyft  & GRU-ODE  & 10    & 24G \\   
		\bottomrule
	\end{tabular}%
 	\caption{Runtime analysis of training cost of different model configs. 'OOM' denotes out-of-memory for one batch in a single A6000 GPU (48G memory)}
\label{table_runtime_analysis}%
\end{table}

\subsection{Baselines}
\textbf{Baselines from prior works} \textbf{Vision track.} FIERY\cite{FIERY} is the first practice for end-to-end stochastic occupancy flow prediction. StretchBEV\cite{StretchBEV}\footnote{StretchBEV-P uses ground-truth labels of past frames as a posterior for prediction, which is unfair for comparison with end-to-end occupancy prediction, so only StretchBEV without labels is compared in the table.} uses a variational autoencoder for learning implicit temporal dynamics and future prediction in a decoupled style. ST-P3\cite{ST-P3} is the first end-to-end planning framework that considers occupancy prediction. BEVerse\cite{BEVerse} is the first multi-task model for both object- and grid-level perception.

\textbf{LiDAR track.} MotionNet\cite{MotionNet} is the first practice for learning BEV grid motion using a simple spatial-temporal voxel-based backbone (STPN). BE-STI\cite{BE-STI} develops the backbone with two new blocks, SeTE and TeSe to enhance temporal feature representation. Since LiDAR track algorithms are not originally proposed for this task, we reimplement them using original BEV backbones and prediction heads for this task.

\textbf{Fusion track.} 
FusionAD\cite{FusionAD} is a multi-modality, multi-task, end-to-end driving framework. They build a transformer to conduct multi-modal BEV-level fusion and downstream perception, prediction and planning tasks. Occupancy prediction results are from the original paper.

\textbf{Baselines of fusion strategies}
\textbf{Synchronous fusion.} The process is first multi-modal spatial fusion, then mixed-modal temporal fusion, and finally standard GRU modules. The assumption is that, at each frame, LiDAR points are tightly synchronized and fused with the nearest images. Spatial fusion follows the same methodology as BEVFusion\cite{BEVFusion}. This process requires strict synchronization and weak time interval uniformity.

\textbf{Asynchronous fusion}: The process is first single-modal temporal fusion, then standard GRU modules, and finally mixed-modal spatial fusion on future timestamps. Single-modal temporal fusion is first performed using the spatio-temporal convolution (STC) unit, and then spatial fusion is performed during prediction. The assumption is that the perception and prediction times are strictly uniform. This process requires only weak synchronization and strict time interval uniformity.

\subsection{Perception Results}
As a similar task to flow prediction, we also compare the performance of BEV segmentation of intermediate representations with prior arts in \cref{tab:comparison_nusc_perception}. We evaluate two main traffic agent categories, vehicle, and pedestrian by IoU metric. With timestamp-agnostic camera-LiDAR fusion by SpatialGRU-ODE, StreamingFlow also achieves impressive progress in typical agent segmentation. It surpasses ST-P3\cite{ST-P3} by $+10.7$ for vehicles and $+22.7$ for pedestrians.

\begin{table}[htbp]
	\centering
	\begin{tabular}{l|cc}
		\toprule
		Method &  Vehicle / IoU & Pedestrian / IoU \\
		\midrule
		VED\cite{VED}  & 23.3  & 11.9 \\
            VPN\cite{VPN}  & 28.2  & 10.3  \\ 
            PON\cite{PON}  & 27.9  & 13.9 \\
            Lift-Splat\cite{LSS} & 31.2 & 15.0 \\
            IVMP\cite{IVMP} & 34.0 & 17.4 \\
		FIERY\cite{FIERY}  & 38.0  & 17.2 \\
            ST-P3 \cite{ST-P3} & 40.1  & 14.5 \\
		\midrule
  		StreamingFlow  & \textbf{50.8}  & \textbf{37.2} \\
		\bottomrule
	\end{tabular}%
 	\caption{Comparison with the state-of-the-art methods for BEV segmentation of vehicles and pedestrians on nuScenes\cite{nuScenes} validation set.}
	\label{tab:comparison_nusc_perception}%
\end{table}

\subsection{Analysis and Discussion}
We provide an intuitive analysis of streaming forecasting training and inference efficiency. Either dense or sparse labels are applicable for the supervised signal of future frames, but the extremely sparse supervised signal as supervision may degrade the performance of SpatialGRU-ODE, as the state may be updated too many times until the next supervised frame. In contrast, the denser supervised signal can strengthen the method to surpass the state-of-the-art synchronized spatial fusion method. For accurate inference at future timestamps with low latency, SpatialGRU-ODE with variable time step, which is closely related to different prediction requests, is the best trade-off for accuracy and latency. 

\subsection{More visualizations}
\label{sec:more_vis}
Demo videos for standard occupancy forecasting, streaming occupancy forecasting and long-term zero-shot or supervised occupancy forecasting for more scenarios will be available at \url{https://github.com/synsin0/StreamingFlow}.

In the 40-frame prediction visualization, the static instances remain static overall with only minor changes in the perception range. Though grid-centric perception is discrete in essence, the prediction results show that StreamingFlow successfully learns the temporal dynamics in continuous time series.
\begin{figure*}
%\begin{figure*}[ht]
	\centering
	\includegraphics[width=0.85 \textwidth]{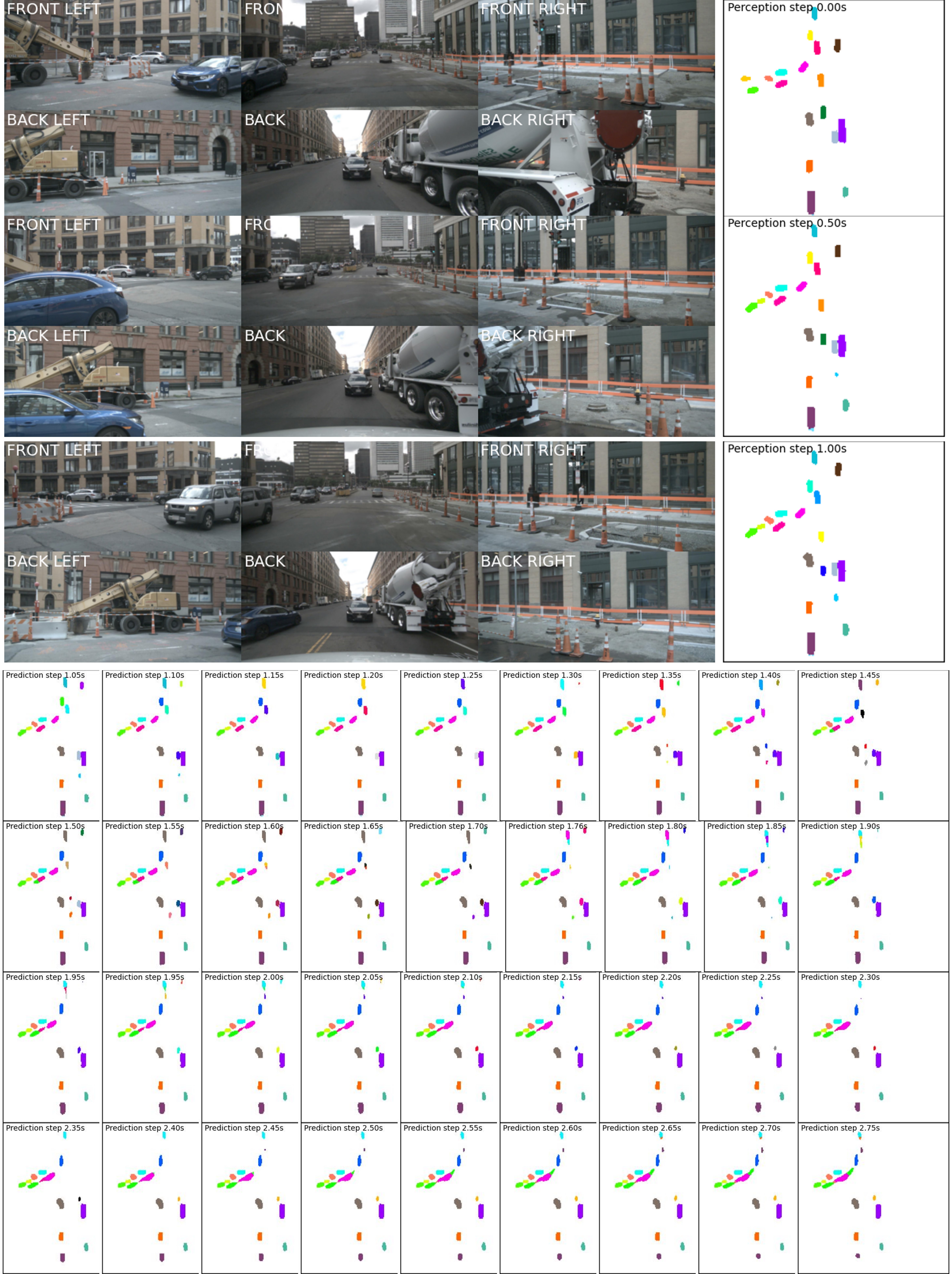}
	\caption{Visualization of StreamingFlow for continuous panoptic occupancy flow prediction at a busy intersection. Given 3 key-frame cameras inputs and asynchronous 5 key-frame LiDAR inputs, we are able to accurately predict the continuous trend of dynamic trend(front left view) only with sparse supervision.}
	\label{visualization}
\end{figure*}

\end{document}